\documentclass[conference]{IEEEtran}
\IEEEoverridecommandlockouts
\usepackage{cite}
\usepackage{amsmath,amssymb,amsfonts}
\usepackage{algorithmic}
\usepackage{graphicx}
\usepackage{textcomp}
\usepackage{xcolor}
\usepackage{multirow}
\usepackage{booktabs,ragged2e}
\usepackage{multicol}
\usepackage[flushleft]{threeparttable}
\def\BibTeX{{\rm B\kern-.05em{\sc i\kern-.025em b}\kern-.08em
    T\kern-.1667em\lower.7ex\hbox{E}\kern-.125emX}}

\begin{document}
\pdfinclusioncopyfonts=1

\title{Improving Text Generation on Images with Synthetic Captions\\
}

\author{\IEEEauthorblockN{1\textsuperscript{st} Junyoung Koh}
\IEEEauthorblockA{\textit{Yonsei University} \\
\textit{ONOMA AI}\\
Seoul, South Korea \\
solbon1212@onomaai.com}
\and
\IEEEauthorblockN{1\textsuperscript{st} Sanghyun Park}
\IEEEauthorblockA{\textit{Yonsei University} \\
\textit{ONOMA AI}\\
Seoul, South Korea \\
aria1th@onomaai.com}
\and
\IEEEauthorblockN{1\textsuperscript{st} Joy Song}
\IEEEauthorblockA{% \textit{} \\
\textit{ONOMA AI}\\
Seoul, South Korea \\
joy.song@onomaai.com}
}
% \author{
% \IEEEauthorblockN{1\textsuperscript{st} Given Name Surname}
%  \and
%  \and
% \IEEEauthorblockN{3\textsuperscript{th} Given Name Surname}

% }
\maketitle

\begin{abstract}
The recent emergence of latent diffusion models such as SDXL\cite{SDXL} and SD 1.5\cite{SD1.5} has shown significant capability in generating highly detailed and realistic images. Despite their remarkable ability to produce images, generating accurate text within images still remains a challenging task. In this paper, we examine the validity of fine-tuning approaches in generating legible text within the image. We propose a low-cost approach by leveraging SDXL without any time-consuming training on large-scale datasets. The proposed strategy employs a fine-tuning technique that examines the effects of data refinement levels and synthetic captions. Moreover, our results demonstrate how our small scale fine-tuning approach can improve the accuracy of text generation in different scenarios without the need of additional multimodal encoders. Our experiments show that with the addition of random letters to our raw dataset, our model's performance improves in producing well-formed visual text.

\end{abstract}


\section{Introduction}
Recent advancements in stable diffusion have revolutionized text-to-image generation. Additionally, along with the improvements in VLMs, there has been a notable improvement in understanding text within images. However, while text-to-image diffusion models have consistently shown exceptional capabilities in generating high-quality images from text prompts\cite{DiffusionBeatsGAN}, they still lack the ability of rendering accurate text within generated images. When such models attempt to mimic text presented in images, such as signs or logos, they often produce nonsensical outputs, referred to as gibberish text. While large-scale models like SDXL may inherently possess some degree of text imitation ability due to biases in datasets like LAION \cite{LAION}, with additional fine-tuning stages, such models are more prone to losing this capability.

Recently, SD3, the latest version of the Stable Diffusion models, demonstrates superior performance in all of the following areas: prompt following, visual aesthetics, and typography \cite{SD3}. In particular, SD3’s ability to generate text outputs within generated images has shown very promising results. 

\begin{figure}[ht]
    \includegraphics[scale=0.29]{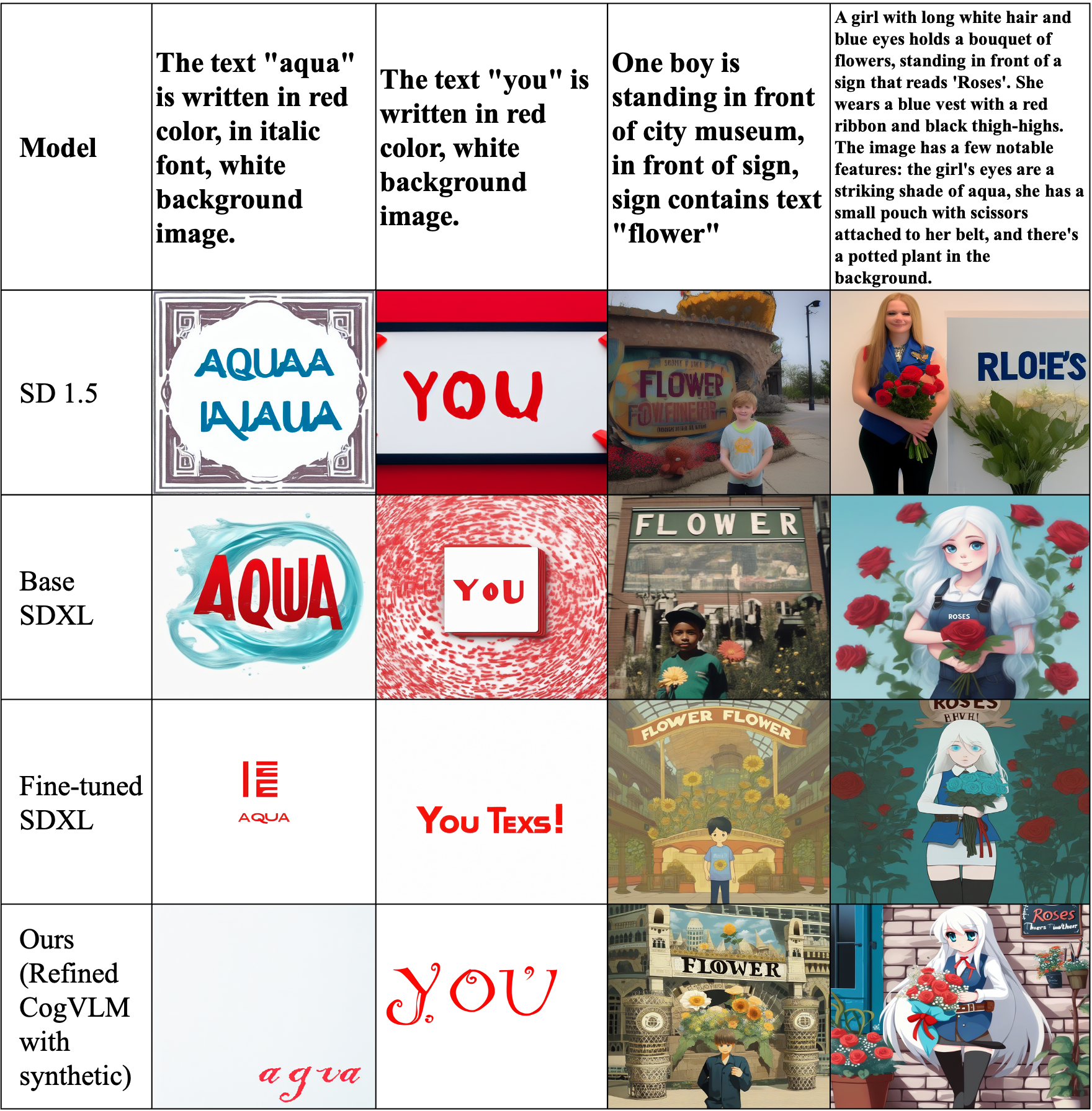}
    \caption{{Comparison of visual results of our model and different text-to-image models. The best results from 16 generated images were selected.
}}
\vspace*{-5mm}
    \label{fig:comparison}
\end{figure}

Although SDXL shows enhanced improvements in creating legible text compared to previous versions, these text generated are not always accurate, often resulting in gibberish text. We conjecture that one such factor could the absence of synthetic captions. Recent works \cite{benefit2, benefit1} have explored the efficiency of synthetic data to augment existing datasets to enhance the performance of intended tasks. Based on prior works, we conducted experiments to examine the extent of which the implementation of synthetic data along with sophisticated data orchestration is applicable to SDXL model. Our contribution can be summarized as follows: (1) Original and generated data orchestration helps enhance the performance of the model. (2) We demonstrate that the loss in text generation ability of SDXL can be remedied through less-intensive fine-tuning techniques and simple dataset curation, consisting of images that contain random characters. (3) We measure the impact of using synthetic captions on generating text displayed in an image by experimenting with varying increments of data and different types of synthetic data (random and detailed).

\begin{figure*}[htbp]
    \includegraphics[width=\textwidth]{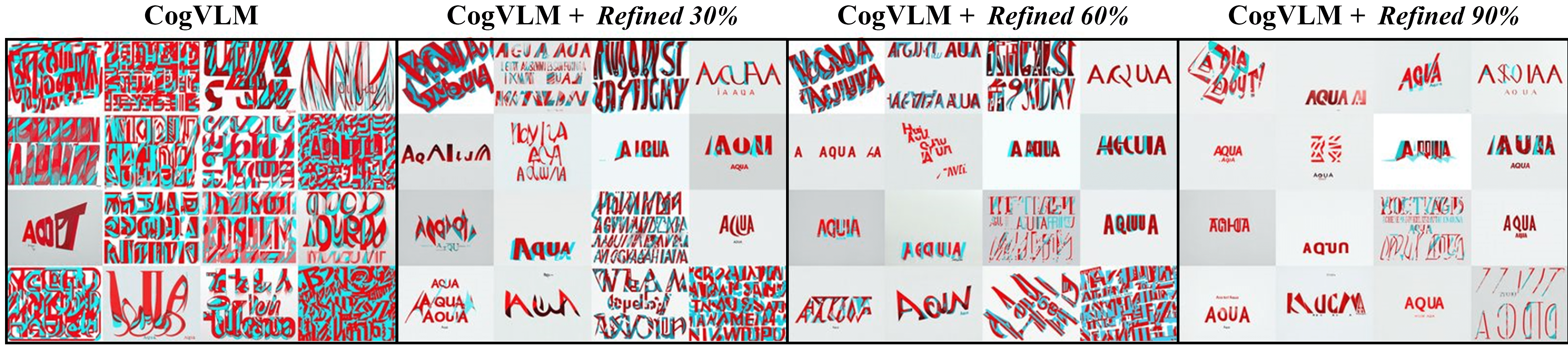}
    \caption{{Comparison of the performance of our model trained on various blending ratios of refined captions and automatic captions}. Only the model trained on CogVLM captioning dataset fails to reconstruct the `aqua' text, whereas models trained with higher percentage of refined data render text more accurately.}
    \label{fig:cogvlmaqua}
\end{figure*}

\section{Related Work}
\subsection{Text-to-image generation}
Recent progress of Text-to-Image models, have been dramatically accelerated with Latent Diffusion Models, and Diffusion Transformers\cite{DiT}. Namely, Stable Diffusion and Stable Diffusion XL are based on Latent Diffusion Models, with U-Net \cite{Unet} backbone. Würstchen\cite{Würstchen} compressed latent space of images reduce computational costs for both training and inference by orders of magnitude. Stable Cascade\cite{Würstchen} is built on a Würstchen pipeline comprising three distinct models that allow for a hierarchical compression of images, achieving remarkable outputs while utilizing a highly compressed latent space. PixArt-$\alpha$\cite{Pixartalpha}, a high-quality diffusion transformer text-to-image synthesis model, achieved superior image generation quality through an exceptionally efficient training process. PixArt-$\delta$\cite{Pixartdelta} incorporates LCM\cite{LCM} and ControlNet\cite{ControlNet} into PixArt-$\alpha$ enables fine-grained control. Pixart-$\sum$\cite{Pixartsigma} is capable of directly generating images at 4K resolution. Playground\cite{playground} enhances color and contrast, improving generation across multiple aspect ratios, and human-centric fine details. 

Some models use CLIP\cite{CLIP} and text encoder to condition the text prompt on diffusion model. DALLE-3\cite{DALLE3}, utilizes unCLIP structure to improve caption understanding. 

\subsection{Synthetic data}
Data refinement has been emerged as critical in training large models, including text-to-image model \cite{benefit6}. However, acquiring high quality data can be challenging due to privacy concerns related to real-world data \cite{benefit8}, data scarcity \cite{benefit7}, and several downsides of data labeling, including high-cost and extensive time consumption \cite{benefit9}. To remedy these problems, synthetic data has emerged as a promising, low-cost alternative to manual annotations \cite{benefit4, benefit10}. Furthermore, due to the inherent limitations of Large Language Models (LLMs), synthetic data is necessary in improving the capabilities of language models \cite{benefit3, benefit5}. Following the LLMs' methodology, training on synthetic captions can substantially improve the prompt following abilities of text-to-image models \cite{DALLE3}.

Instead of merging real and synthetic data generated by base text-to-image diffusion models, adding synthetic data from a fine-tuned model can enhance the performance of such models \cite{benefit1}. Unlike methods that enhance the quality of text-to-image generation through the addition of more sophisticated captions, such as ImageNet classification \cite{benefit1}, and DALL-E-3 \cite{DALLE3}, we demonstrate the extent to which generative data is effective with SDXL when rendering accurate text within images.

\begin{table*}[htbp]
\caption{Mixture of Dataset Experiment}
\begin{center}
\begin{tabular}{|cc|c}
\toprule
\multicolumn{1}{c|}{\textbf{Raw Data}}       & \textbf{Synthetic Data}                                         &              \textbf{Total Dataset}                     \\ 
\midrule
\multicolumn{1}{c|}{-}                   & Detail guidance of text and random letter with white background (495) & 495                           \\ 
\multicolumn{1}{c|}{CogVLM (688)}  & -                                                            & 688                               \\ 
\multicolumn{1}{c|}{CogVLM + refined 30\% (688)}  & -                                                            & 688                               \\ 
\multicolumn{1}{c|}{CogVLM + refined 60\% (688)}  & -                                                            & 688                               \\ 
\multicolumn{1}{c|}{CogVLM + refined 90\% (688)}  & -                                                            & 688                               \\ 
\multicolumn{1}{c|}{CogVLM + refined 100\% (688)} & -                                                            & 688                               \\
\multicolumn{1}{c|}{CogVLM + refined 100\% (688)} & Word text with white background  (165)                               & 853                         \\
\multicolumn{1}{c|}{CogVLM + refined 100\% (688)} & Detail guidance of text with white background (165)                   & 853                         \\
\multicolumn{1}{c|}{CogVLM + refined 100\% (688)} & Random letter with white background (165)                   & 853                         \\
\multicolumn{1}{c|}{CogVLM + refined 100\% (688)} & Detail guidance of text and random letter with white background (330) & 1018                   \\
\multicolumn{1}{c|}{CogVLM + refined 100\% (688)} & Random letter with white background (330) & 1018                   \\
\bottomrule

\end{tabular}
\label{tab1}
\end{center}
\end{table*}

\section{Method}

% \ref{AA} -- \ref{SCM}
Our procedure consists of three steps: (1) create training dataset from existent data and synthetic data; (2) measure the impact of synthetic data by experimenting with different mixtures of base data and generated data; (3) train eleven text-to-image models on the dataset with different combinations of base and generated dataset ratios.

\subsection{Data preparation}\label{DataPre}
We conduct experiments based on two types of datasets.
\begin{enumerate}
    \item A sample of 688 images from Danbooru 2023 dataset\cite{danbooru}, which is a large-scale anime illustration dataset with tags ``sign". We use CogVLM\cite{cogVLM} to generate more descriptive, textual captions based on tag information. 
    \item 165 artificial text-image pairs of optical character renderings on white background. 
\end{enumerate}

We investigate two approaches to dynamically refine image-caption pairs.
\begin{itemize}
    \item We re-annotate the captions either manually or automatically to include a more accurate description of text, lowering the inconsistencies between the image and caption. This process can be conducted via GPT-4\cite{GPT4}, Gemini Pro Vision\cite{GeminiPro}, or manual labor. 

\end{itemize}
\begin{itemize}
    \item We scale the density of information for which the caption holds for its corresponding image. The levels are specified as follows: basic, detailed, and random series. Basic information simply describes the text content (i.e. the image with text ‘cat’). Detailed information more specifically depicts the text’s color, thickness, scale, location, and whether or not the text is italicized. Random series refers to the random combination of characters to form new, potentially unseen phrases, which are not expected to be seen in the pretraining stages.
\end{itemize}

\subsection{Mixture of base dataset and synthetic dataset}
\begin{figure*}[ht]
    \includegraphics[width=\textwidth]{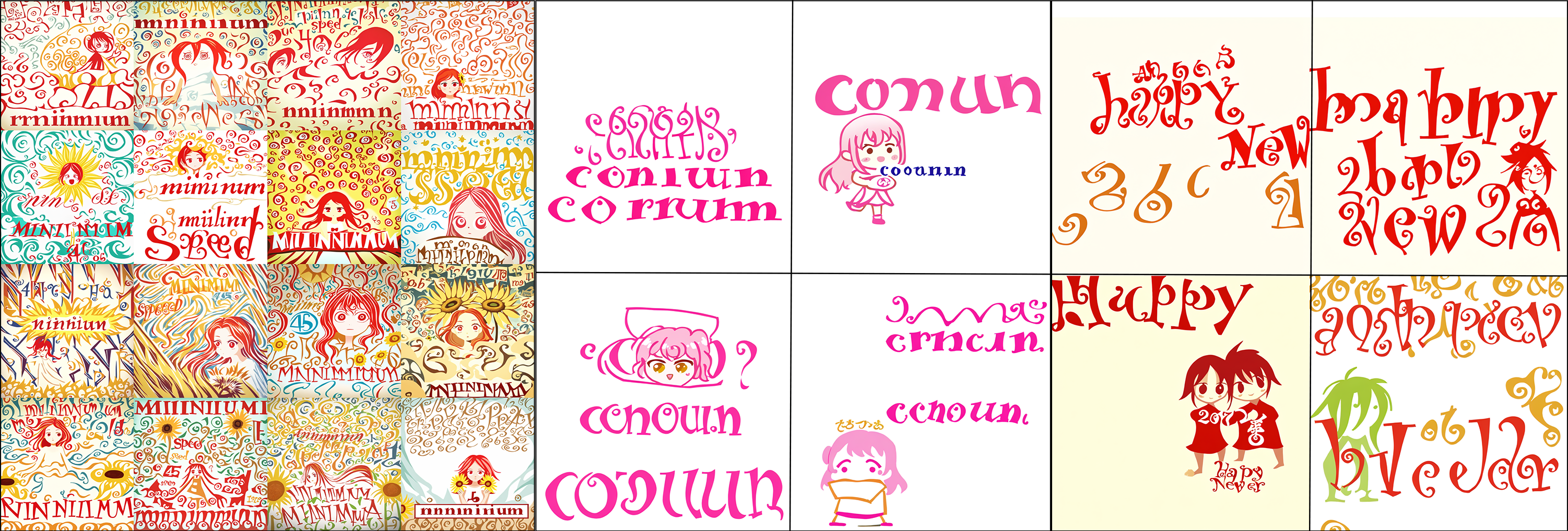}
    \caption{{The inference result of model when synthetic dataset is exclusively used. Model fails to follow the prompt, especially related to background. The prompt used for inference is ``A girl with long red hair and red eyes holds a sunflower amidst a whirlwind of signs, showcasing a dreamy and surreal atmosphere. The sign has a English text with `MINIMUM SPEED 45'.''
``A girl with pink hair and a purple top hat is holding a sign that reads `Conun Drum', with a surprised expression on her face. The image has an artistic and whimsical feel, with the girl's attire and the background design adding to the fantasy theme.
``A young male with black hair and red eyes is holding a phone, surrounded by a dynamic environment with a road sign and a sign. With the text `Happy Birthday IZAYA'.''}}
    \label{fig:modecollapse}
\end{figure*}

To observe the effect of dataset refinement on model performance, we propose to adjust the ratio of manual captions to automatic captions. As the measure of refinement on the raw dataset increases, the model's ability to render accurate text within image also improves at a significant level as shown in Figure \ref{fig:cogvlmaqua}. We also examine that sole synthetic dataset incurs severe catastrophic forgetting, as depicted in Figure \ref{fig:modecollapse}. Thus, to assess the impact that synthetic data has on the existing raw dataset, we added different types of synthetic data depending on the raw data's refinement level. Moreover, we demonstrate that the quantity of synthetic data affects the model's performance. Table \ref{tab1} depicts the details of our dataset composition for each experiment.

\begin{figure}[ht]
    \includegraphics[scale=0.305]{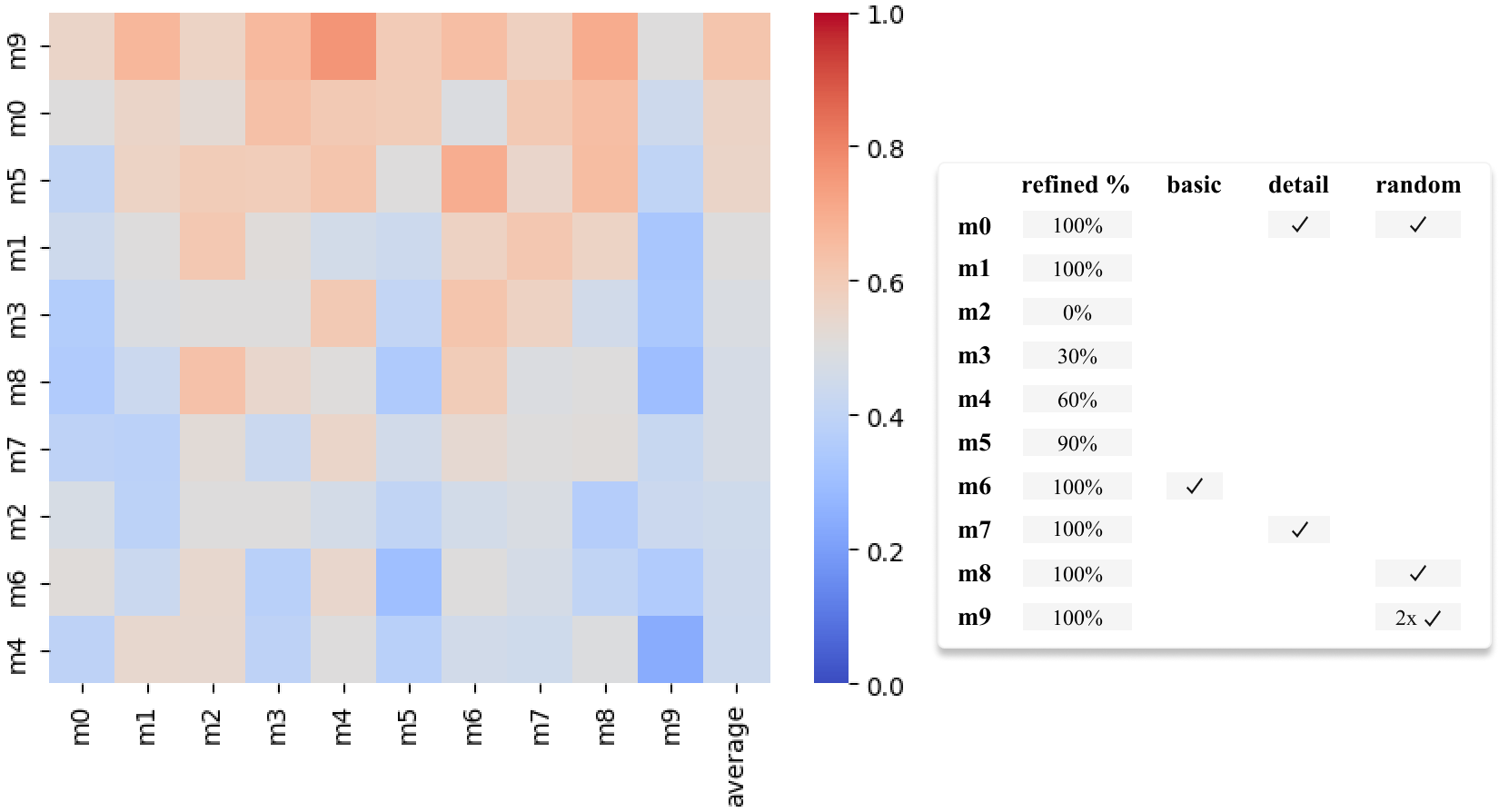}
    \caption{{Heat map comparing the performance of our different models through human evaluation results. We provide a legend (right) that shows the detailed data composition for each model (m0 to m9) in terms of refinement level and type of synthetic data.
}}
\vspace*{-3mm}
    \label{fig:model10comparison}
\end{figure}

\subsection{Model training}
We trained the model using the following hyperparameters. We used U-net learning rate of 3e-6, learning rate for Text Encoder1 6e-7 same as learning rate for Text Encoder2. We used scheduler ``constant'' with warm up steps 50. Our optimizer type is Lion8bit\cite{Lion8bit} trained with 4,165 steps, 20 epoch. We used one A6000, utilizing max 23GB of VRAM in average. Training the model took approximately 140 to 180 minutes.

\begin{figure*}[ht]
    \includegraphics[width=\textwidth]{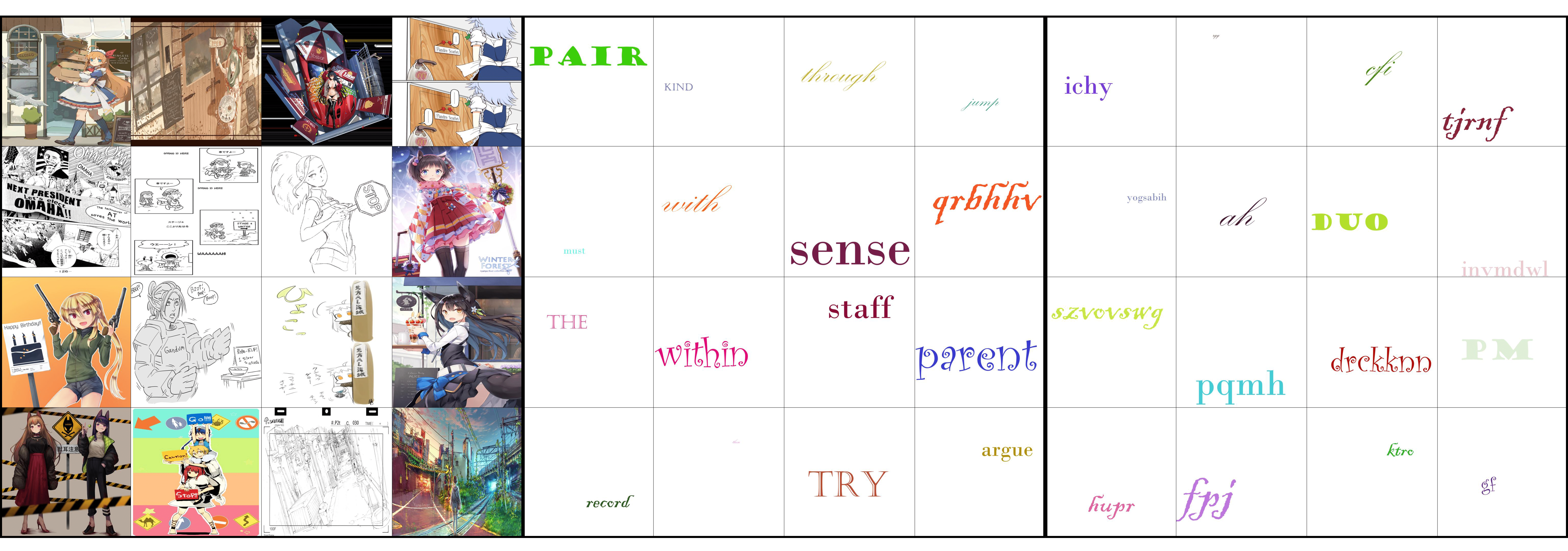}
    \caption{{Dataset consisting of raw data from Danbooru as well as synthetic data.} This image shows sample raw data from Danbooru(left), sample synthetic data for real words (center) and sample synthetic data composed of random combination of letters (right).}
    \label{fig:dataset}
\end{figure*}

\section{Evaluation}

\subsection{Human evaluation}
\begin{figure*}[ht]
\centering
\includegraphics[width=0.9\textwidth]{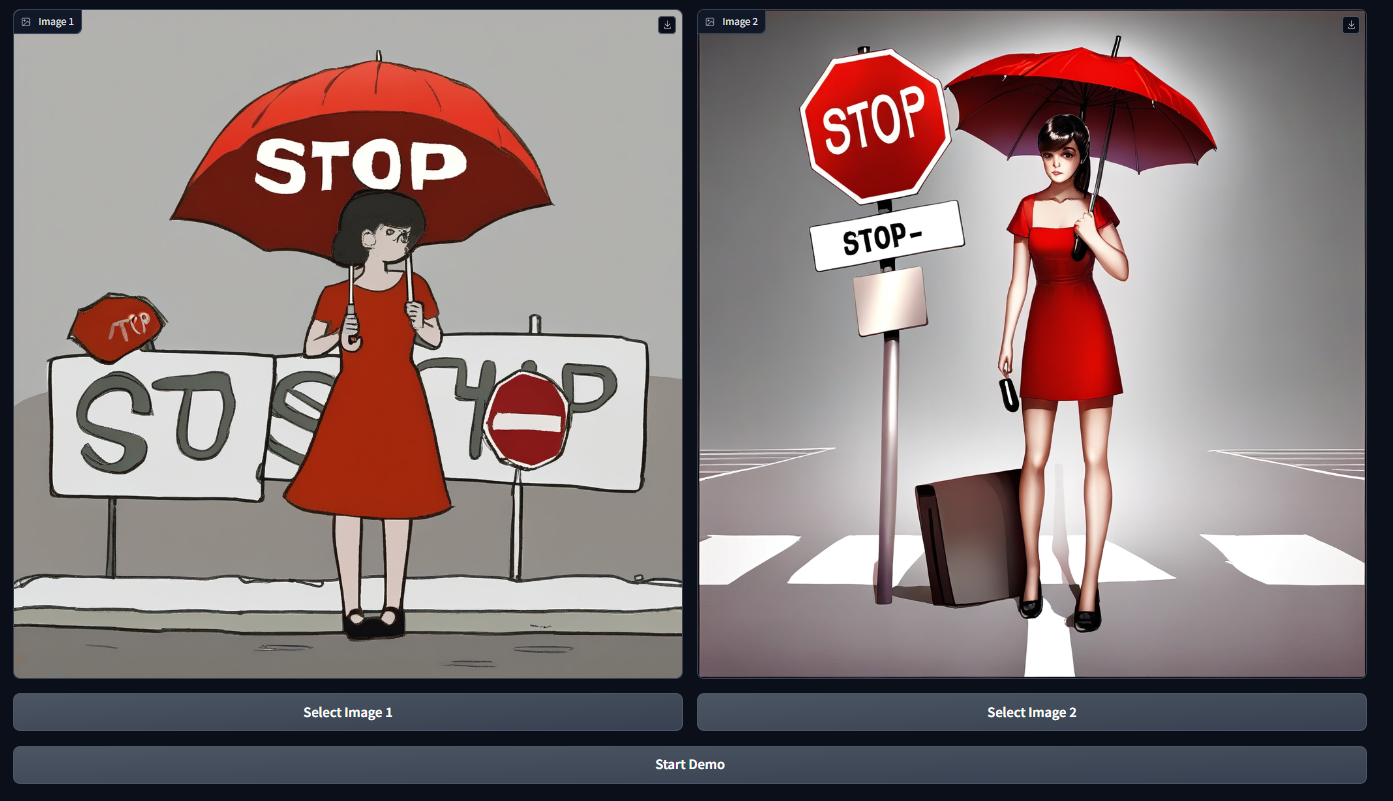}
\caption{{Interface for human evaluation. We present human raters with two side-by-side images that were generated from the same caption and ask raters to choose which image is better at generating more legible text.}} 
\label{fig:humanevaluationprove}
\end{figure*}

\begin{figure*}
\centering
\begin{minipage}[b]{.45\textwidth}
{\includegraphics[width=\textwidth]{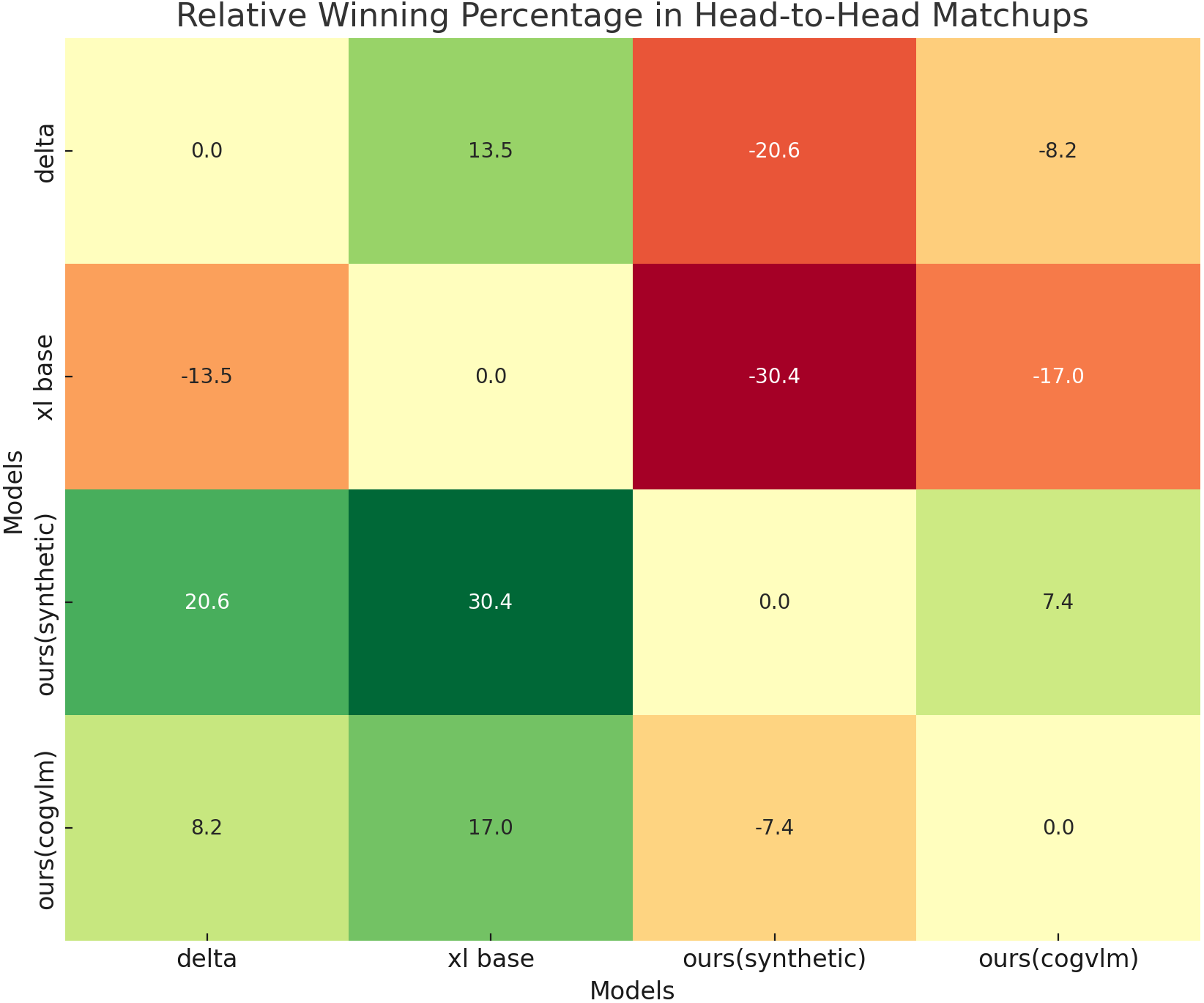}}
\caption{Human evaluation results for our model versus other text-to-image generation models. This figure shows how the images generated by our model are preferred by human raters over other competitor models.}\label{fig:relativeModel}
\end{minipage}\qquad
\begin{minipage}[b]{.45\textwidth}
 {\includegraphics[width=\textwidth]{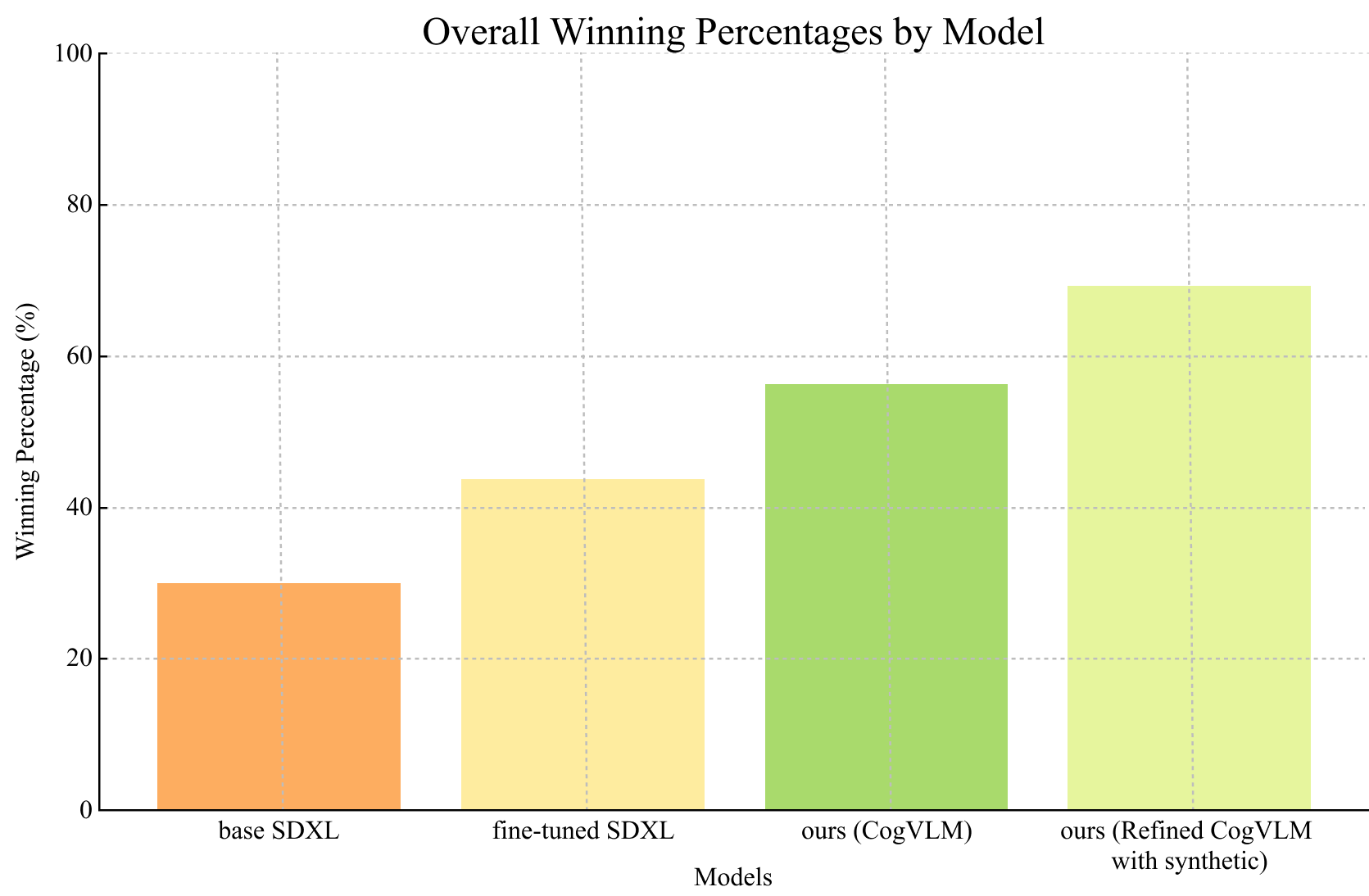}}
\caption{Comparison of winning percentages of different models. Our generative approach (synthetic data combined with base dataset 100\% refined captioned by CogVLM) enables generation of more accurate text within images.}\label{fig:Overall winning}
\end{minipage}
\end{figure*}

We conducted a human evaluation deployed into gradio WebUI as shown in Figure \ref{fig:humanevaluationprove}. We collected and evaluated 2788 responses for 1450 generated samples. The performance results of our eleven models are visualized in Figure \ref{fig:model10comparison}.
Figure \ref{fig:relativeModel} and Figure \ref{fig:Overall winning} show how our model m0, composed of 100\% refined raw data with detail and random synthetic data, outperforms other base or fine-tuned models. The integration of synthetic data notably enhances the model's ability in generating text.

\begin{table*}[htbp]
\begin{center}
\begin{threeparttable}
\caption{Metric Comparison}
\setlength\tabcolsep{8pt}
\begin{tabular}{lcccccccccccc}
\toprule
    \multirow{2}{*}{\begin{tabular}[c]{@{}l@{}}Model's\\ Metric\end{tabular}} &
    \multicolumn{3}{c}{SD 1.5} &
    \multicolumn{3}{c}{Base SDXL} &
    \multicolumn{3}{c}{\begin{tabular}[c]{@{}c@{}}Fine-tuned \\ SDXL\tnote{d}\end{tabular}} &
    \multicolumn{3}{c}{Ours} \\ 

       & CER\tnote{a}  & CLIP\tnote{b} & TIFA\tnote{c}  & CER    & CLIP   & TIFA  & CER  & CLIP &TIFA  & CER  & CLIP & TIFA \\
       
    \midrule
flower & 0.5442  & 0.3490 & 0.7533 & 0.5394 & \textbf{0.3609} & \textbf{0.7788} & 0.4514 & 0.2899 & 0.7773 & \textbf{0.1938} & 0.3151 & 0.7347 \\
roses  & 0.9833  & 0.3194 & 0.4888 & 0.8856 & 0.3270   & 0.6341 & 0.4667 & 0.3448 & 0.7856 & \textbf{0.2872} & \textbf{0.3536} & \textbf{0.8717} \\
aqua   & \textbf{0.5800} &  0.2839  & 0.4509 & 0.5864 & 0.3019   & 0.5402 & 0.6792 & o.2956   & 0.7262 & 0.7479  & \textbf{0.3195}  & \textbf{0.8337} \\
you    & 0.6719 & 0.3422     & 0.4241 & 0.6753 & 0.3633   & 0.7231 & 0.7892 & 0.3359     & 0.8571 & \textbf{0.3500} & \textbf{0.3711}   & \textbf{0.9040} \\
    \bottomrule
\end{tabular}

\label{tab2}
\begin{tablenotes}
\RaggedRight
\item[a] CER : Character Error Rate, Compared original text with OCR detection text from Google Cloud Vision
\item[b] CLIP : Compared original prompt with generated image using ViT-B/32
\item[c] TIFA : Compared VQA model Answer with LLM Question-Answer sheet from GPT-4-Vision
\item[d] Kohaku-XL Delta
\end{tablenotes}
\end{threeparttable}
\end{center}
\end{table*}

\begin{figure}[htbp]
    \includegraphics[scale=0.12]{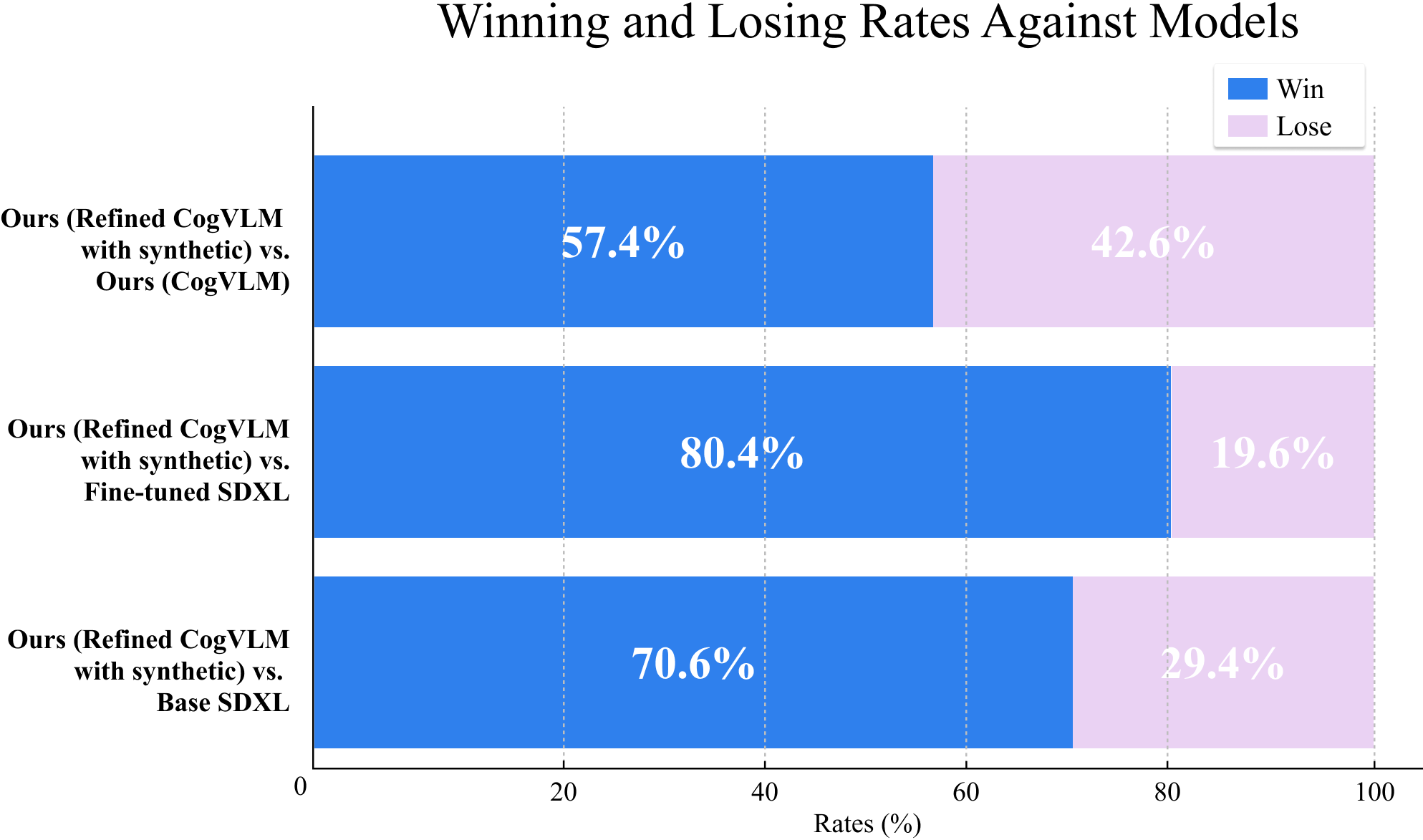}
    \caption{{Human evaluation results, where we show the win vs. lose percentages of our model (refined CogVLM with synthetic data). Our model is preferred by human annotators than the base SDXL, fine-tuned SDXL, and our other model (CogVLM) in terms of text rendering.}}
\vspace*{-3mm}
    \label{fig:winning against}
\end{figure}

\subsection{CER metrics}
Optical Character Recognition (OCR) has been extensively studied in the academia field. OCR is a technology that detects and extracts text images into plain text\cite{OCR1, OCR2, OCR3}. Here, to measure the performance, we used Google Cloud Vision, with its robust OCR detection capabilities and rapid processing speed, ensuring high recognition rates.  We compare the model's performance by utilizing the Character Error Rate (CER) metrics to evaluate the OCR quality output. The results, presented in Table \ref{tab2}, show that our model has better generation capacity than the other models. However, we were not able to perform tests on diverse fonts, due to OCR performance limitations.

\subsection{CLIP score}
To measure the prompt fidelity, we used OpenAI-VitB/32 model to calculate the CLIP score, comparing the similarity score between the target prompt and generated images. The prompts, which have more than 77 tokens, has been padded into multiple chunks, and batched to calculate the mean CLIP embedding vector. Our results, presented in Table \ref{tab2}, show that our fine-tuning approach improves prompt fidelity, recovering natural language processing ability of original SDXL in some aspects. However, it is notable that CLIP score is not suitable for animation  styled images or text-retrieval tasks, resulting in an inconsistent CLIP score.

\subsection{TIFA score}
\begin{figure*}[ht]
    \centering
    \includegraphics[scale=0.11]{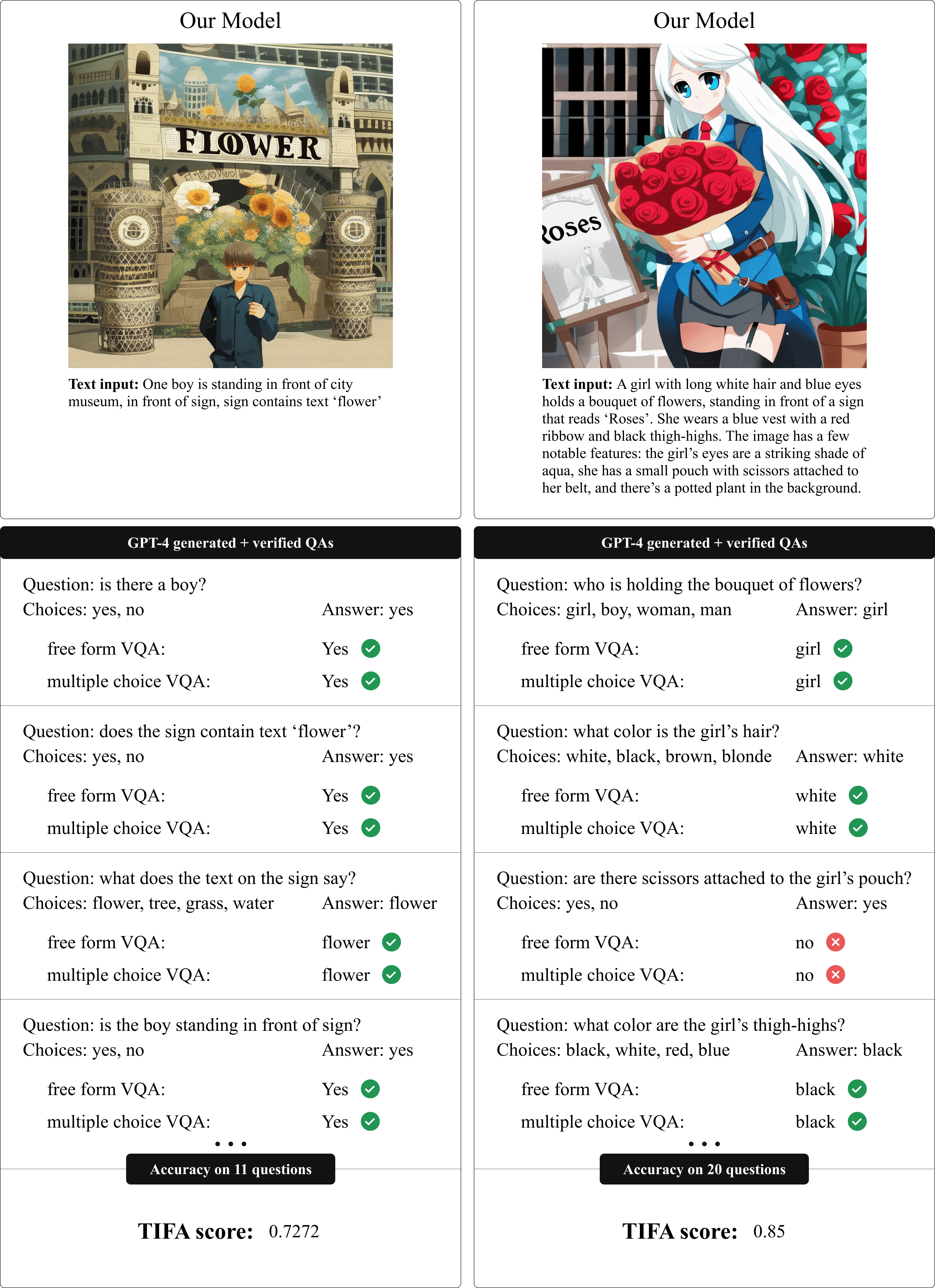}
    \caption{{Given the text prompt, TIFA uses GPT-4 to generate several question-answer pairs and a QA model filters them. The text input on the left (4 out of 11 questions are displayed). The text input on the right (4 out of 20 questions are presented). Our high TIFA scores for both text inputs demonstrate that the VQA models can accurately answer the questions given our model's generated image.}}
    \label{fig:TIFA}
\end{figure*}

TIFA\cite{tifa} measures the faithfulness of a generated image to its text input via visual question answering (VQA). We used language model as OpenAI GPT-4-Vision and VQA model as mPLUG-large \cite{mplug}. Given the text prompt, TIFA uses GPT-4 to generate several question-answer pairs and filtered by UnifiedQA-v2-t5-large-1363200. Our results, presented in Table \ref{tab2}, show that VQA model captures more various aspect for our fine-tuning approach. The sample generated question and answer pair is visualized on Figure \ref{fig:TIFA}.

\begin{figure}[ht]
\includegraphics[width=0.5\textwidth]{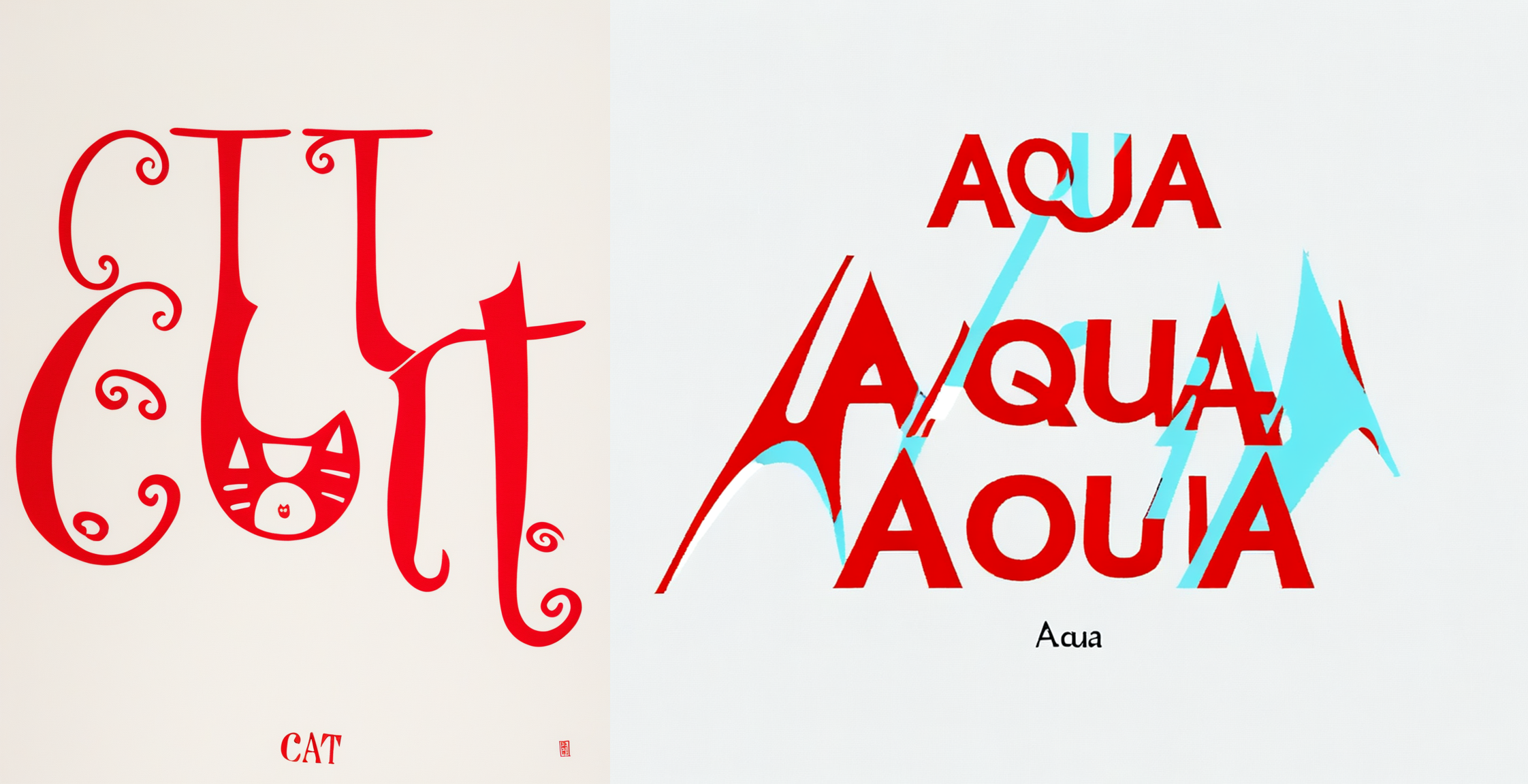}
\caption{{Inference result of ``the text `cat' is written in red color, in white background", ``the text `aqua' is written in red color, white background". The model which is exclusively trained on synthetic dataset, suffers separating text from concept generation.}}
\label{fig:semanticleakage}
\end{figure}

\section{Discussion}
To examine the effect of random series dataset, we train the models based on refined CogVLM captions and increased synthetic dataset as shown in Table \ref{tab1}. Figure \ref{fig:model10comparison} shows that m9 model outperforms all other models, including those that were trained on the mixture dataset. This implies the following: First, the refinement level is a crucial aspect in model training. We observed that models trained on lower refinement levels demonstrated inferior performance compared to those trained on high refinement levels, such as m2, highlighting the relationship between the amount of refined data and model accuracy.
Second, expanding the dataset with monotonous, synthetic image-caption pairs that contain only the content of the text proved to be harmful to the model performance. Furthermore, the addition of datasets comprised of random characters or those featuring detailed textual descriptions does not yield a substantial impact when applied to small-scale datasets. In contrast, when dealing with large datasets, the supplementation of random characters proves to be effective. Lastly, the model's ability to generate accurate visual text significantly diminishes when solely relying on synthetic data.

\section{Conclusion}
With the growing interest in diffusion models and their impressive generation capabilities, it has become essential to explore their ability to generate clear and coherent textual content within images. As such, we focused on analyzing the effectiveness of fine-tuning in aspect of improving text rendering abilities. Overall, our findings show that extensive refinement of datasets is vital for enhancing the model's performance. Thoughtful curation of datasets is crucial, particularly to prevent model knowledge drift. The incorporation of our random synthetic dataset effectively mitigates the knowledge drift phenomenon, enabling high task performance. However, we encountered the following several limitations and will plan to overcome such limitations in the follow-up study:
\subsection{Occurrence of mode collapse}\label{BB}
As explained in Figure \ref{fig:modecollapse}, we observed that the overly high ratio of synthetic data lacking diversity incurs mode collapse. This over-fitting behavior is common when the training dataset is limited \cite{mode collapse}. In order to analyze how the application of data curation and refinement impacts model performance, we conducted an initial experiment considering the most simple scenario. For future works, we will further investigate the effect of various styled synthetic dataset on boosting the results.

\subsection{Semantic leakage}
Semantic leakage, one of the limitations of text-to-image generation models, refers to the incorrect associations between entities and their visual attributes \cite{semantic leakage1, semantic leakage2}. Here, we validate that the model fails to resolve the bound attributes, especially in ``text'' and ``description'', as depicted in Figure \ref{fig:semanticleakage}.  Several studies\cite{semantic leakage3, semantic leakage4} show that synthetic, dense captions are vital for model performance, solving attribute leakages. Combining these insights, we conclude that the composition of various datasets, application of refined captions, and utilization of both synthetic and real data are critical factors in alleviating semantic leakage.

\subsection{Zero-shot capability on random text pair}
Diffusion models often suffer from spelling inaccuracies when rendering text within the generated images\cite{limitation3-2, limitation3-3}. We discovered that our model fails more frequently at generating accurate and clear textual representations within images when generating random characters, rather than meaningful words. This diminishing performance can be potentially derived from two primary factors. First, the model itself may have been heavily trained on LAION dataset, mainly in logos and signs. In addition, the instability of CLIP text encoder in aspect of handling characters may lead to sub-optimal performance in calculating embedding similarities\cite{limitation3-1}. Such limitations has lead the recent trend of utilizing large scale multi-modal encoders in diffusion models, an approach that we can take in the future.

\section{Acknowledgements}

This work was partly supported by an IITP grant funded by the Korean Government
(MSIT) (No. RS-2020-II201361, Artificial Intelligence Graduate School Program (Yonsei
University)).

\vspace{10pt}

% \section*{References}

% \vspace{12pt}
% \color{red}
% IEEE conference templates contain guidance text for composing and formatting conference papers. Please ensure that all template text is removed from your conference paper prior to submission to the conference. Failure to remove the template text from your paper may result in your paper not being published.

\newpage
\onecolumn
\section*{Appendix}

\vspace*{-4cm}
\begin{figure*}[ht]
\includegraphics[width=\textwidth]{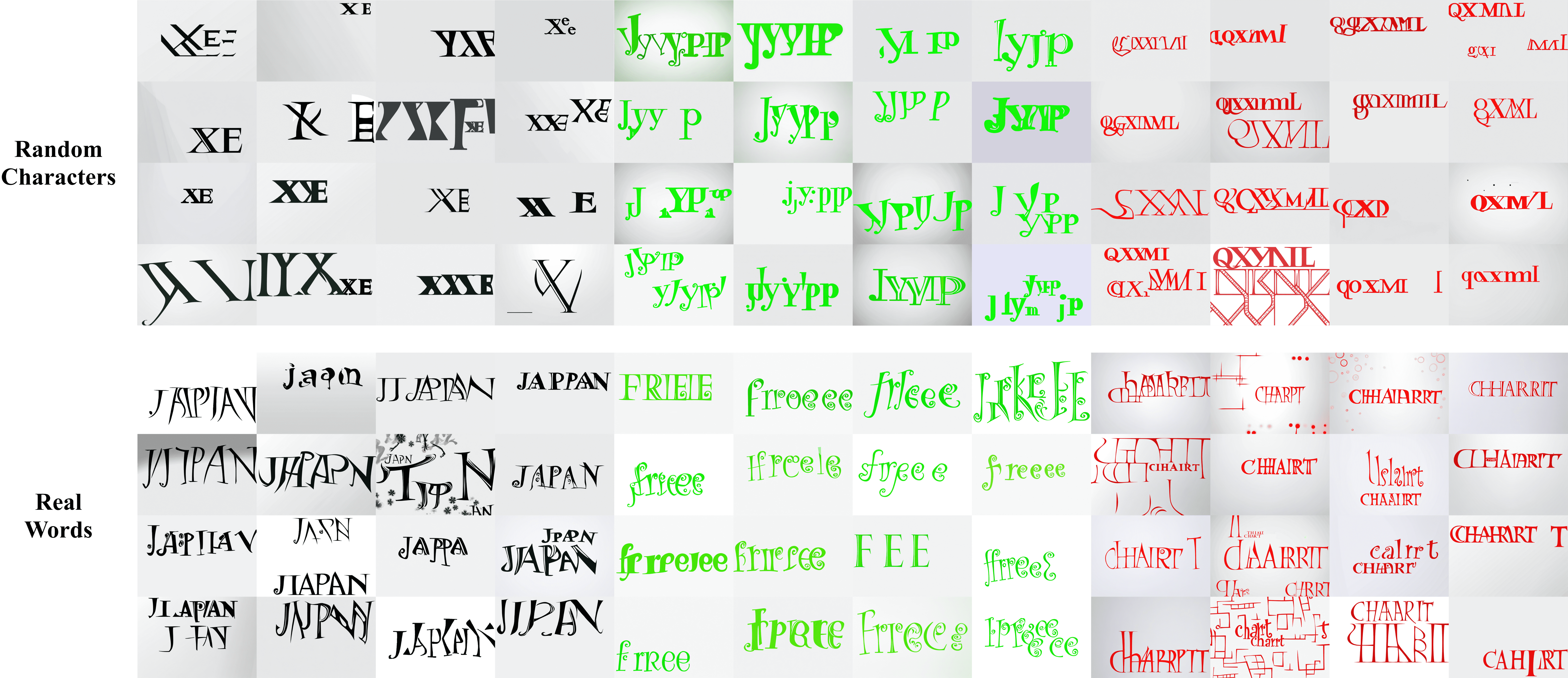}
\caption{{Comparison in visual results of our model's performance in generating random characters and existing characters. Our model fails more frequently at generating accurate and clear textual representations within images for random characters than for existing words. Prompts (top) are: (1) \textit{The image contains text ``xe" written in black color in white background}, (2) \textit{The image contains text ``jyp" written in lime color in white background}, (3) \textit{The image contains text ``qxml" written in red color in white background}. Prompts (bottom) are: (1) \textit{The image contains text ``japan" written in black color in white background}, (2) \textit{The image contains text ``free" written in lime color in white background}, (3) \textit{The image contains text ``chart" written in red color in white background.}}}
     \label{fig:zeroshot}
\end{figure*}

\end{document}